\documentclass[letterpaper]{article} % DO NOT CHANGE THIS
\usepackage{aaai2026}  % DO NOT CHANGE THIS
\usepackage{times}  % DO NOT CHANGE THIS
\usepackage{helvet}  % DO NOT CHANGE THIS
\usepackage{courier}  % DO NOT CHANGE THIS
\usepackage[hyphens]{url}  % DO NOT CHANGE THIS
\usepackage{graphicx} % DO NOT CHANGE THIS
\usepackage{natbib}  % DO NOT CHANGE THIS AND DO NOT ADD ANY OPTIONS TO IT
\usepackage{caption} % DO NOT CHANGE THIS AND DO NOT ADD ANY OPTIONS TO IT
\usepackage{algorithm}
\usepackage{algorithmic}

\usepackage{newfloat}
\usepackage{listings}
\DeclareCaptionStyle{ruled}{labelfont=normalfont,labelsep=colon,strut=off} % DO NOT CHANGE THIS
\floatstyle{ruled}
\newfloat{listing}{tb}{lst}{}
\floatname{listing}{Listing}
\usepackage{booktabs}
\usepackage{amsmath}
\usepackage{xcolor}
\usepackage{multirow}
\usepackage{siunitx}
\usepackage{amssymb}
\usepackage{graphicx}
\usepackage{bm}

\title{When Genes Speak: A Semantic-Guided Framework for Spatially Resolved Transcriptomics Data Clustering}

\author{
    Jiangkai Long\textsuperscript{\rm 1},
    Yanran Zhu\textsuperscript{\rm 1},
    Chang Tang\textsuperscript{\rm 2}\thanks{Corresponding author}
    Kun Sun\textsuperscript{\rm 1},
    Yuanyuan Liu\textsuperscript{\rm 1},
    Xuesong Yan\textsuperscript{\rm 1}
}

\affiliations{
    \textsuperscript{\rm 1}School of Computer Science, China University of Geosciences, Wuhan 430074, China\\
    \textsuperscript{\rm 2}School of Software Engineering, Huazhong University of Science and Technology, Wuhan 430074, China\\
    longjk@cug.edu.cn, tangchang@hust.edu.cn
}

\usepackage{bibentry}
\begin{document}

\maketitle

\begin{abstract}
Spatial transcriptomics enables gene expression profiling with spatial context, offering unprecedented insights into the tissue microenvironment. However, most computational models treat genes as isolated numerical features, ignoring the rich biological semantics encoded in their symbols. This prevents a truly deep understanding of critical biological characteristics.
To overcome this limitation, we present SemST, a semantic-guided deep learning framework for spatial transcriptomics data clustering. SemST leverages Large Language Models (LLMs) to enable genes to “speak” through their symbolic meanings, transforming gene sets within each tissue spot into biologically informed embeddings. These embeddings are then fused with the spatial neighborhood relationships captured by Graph Neural Networks (GNNs), achieving a coherent integration of biological function and spatial structure.
We further introduce the Fine-grained Semantic Modulation (FSM) module to optimally exploit these biological priors. The FSM module learns spot-specific affine transformations that empower the semantic embeddings to perform an element-wise calibration of the spatial features, thus dynamically injecting high-order biological knowledge into the spatial context.
Extensive experiments on public spatial transcriptomics datasets show that SemST achieves state-of-the-art clustering performance. Crucially, the FSM module exhibits plug-and-play versatility, consistently improving the performance when integrated into other baseline methods.
\end{abstract}

\begin{links}
    \link{Code}{https://github.com/longjiangk/SemST}
\end{links}

\section{Introduction}
Understanding the intricate spatial architecture of tissues is a cornerstone of modern biology and medicine, holding the key to deciphering mechanisms of organogenesis, disease progression, and therapeutic response \cite{staahl2016visualization, longo2021integrating}. The advent of spatial transcriptomics has revolutionized this pursuit by enabling the measurement of gene expression profiles while preserving their native spatial coordinates \cite{rodriques2019slide, marx2021method}. A fundamental computational task in spatial transcriptomics is the identification of spatial domains—regions with similar gene expression patterns that often reflect an underlying anatomical or functional organization \cite{biancalani2021deep}. Robust spatial domain identification lays a solid foundation for downstream biological analyses, such as differential gene expression analysis, functional enrichment analyses, and cell–cell interaction inference \cite{dong2022deciphering}. This task is commonly formulated as a clustering problem, aiming to partition the tissue into biologically meaningful units \cite{du2023advances}. 

Recent years have seen a surge in deep learning models, particularly those based on Graph Neural Networks (GNNs) \cite{li2022cell, liu2024comprehensive, zahedi2024deep}. These methods have achieved considerable success by modeling the tissue as a graph, where spots (or cells) are nodes and spatial proximity defines the edges. By aggregating information from neighboring spots, GNNs can effectively learn representations that capture the spatial context of gene expression. Another advantage is that the graph topology naturally reflects possible cell–cell interactions, offering an implicit yet biologically meaningful view of intercellular communication \cite{zhou2022graphing, lampert2024cell}. 

However, a critical and pervasive limitation persists: these models overwhelmingly treat genes as isolated, high-dimensional numerical features. This approach completely ignores the vast repository of biological knowledge encapsulated in the gene symbols themselves—knowledge about their functions, pathways, and interactions. For example, “Postn” has been implicated in extracellular matrix remodeling and fibroblast activation, whereas the co-expression of “Ttn” and “Actc1” is indicative of cardiac muscle identity. This semantic gap prevents the models from reasoning about the tissue in a biologically informed manner, thereby limiting their capacity for deeper biological discovery.

The recent success of general-purpose Large Language Models (LLMs) in bioinformatics \cite{liu2024large} has opened new possibilities for leveraging textual gene knowledge in computational biology. For instance, GenePT \cite{chen2025simple} leverages ChatGPT \cite{brown2020language} to generate embeddings for individual genes based on their textual descriptions or symbols, facilitating diverse applications in gene and single-cell analysis. In the context of integrating transcriptomics with histopathology, models such as SGN \cite{yang2024spatial} and OmiCLIP \cite{chen2025visual} adopt a similar approach, generating gene semantic embeddings alongside image embeddings, which enables joint modeling of gene symbols and tissue morphology. These works demonstrate the feasibility of translating gene symbols into biologically meaningful concepts. However, despite its promise, the direct application of this semantic-aware approach to spatial transcriptomics data clustering has remained challenging and largely unexplored. The primary challenge lies in the fundamental disparity between symbolic biological knowledge and quantitative spatial information, two modalities with inherently different characteristics that are difficult to reconcile.

To bridge this gap, we introduce SemST, a novel semantic-guided framework for spatial transcriptomics data clustering. SemST explores the integration of gene semantics directly into the spatial modeling process. We leverage the power of LLMs, pre-trained on vast biomedical text corpora, to transform discrete gene symbols into dense, continuous embeddings that map spots within a meaningful biological semantic space. The central challenge  then becomes how to effectively integrate these powerful semantic priors with the spatial relationship features. A naive fusion risks diluting the rich information from each modality, rather than creating a synergistic whole. 

To this end, we introduce the Fine-grained Semantic Modulation (FSM) module. Instead of passively combining information, the FSM module enables an active, dynamic “dialogue” between biological functions and spatial contexts. Specifically, the FSM learns a spot-specific affine transformation to perform an element-wise calibration of the GNN-derived spatial features. This mechanism allows high-order biological knowledge to dynamically guide and refine the representation of local spatial context, leading to more robust and biologically plausible representations.

The contributions of this work can be summarized as the following three-fold: 
\begin{itemize}
\item We propose SemST, a novel deep learning framework that effectively fuses LLM-derived gene semantics with GNN-captured spatial context for superior spatial domain identification, moving beyond the conventional treatment of genes as isolated numerical features.

\item We design the Fine-grained Semantic Modulation (FSM) module, which enables semantic priors to dynamically calibrate spatial features in an element-wise manner, resulting in the infusion of high-order biological knowledge into spatial representations.

\item Extensive experiments on multiple benchmark spatial transcriptomics datasets demonstrate that SemST significantly outperforms existing state-of-the-art methods. We also show that our proposed module is a versatile, plug-and-play component that consistently boosts the performance of other baseline models.
\end{itemize}

\section{Related Work}

\subsection{Large Language Models for Biological Semantics}

Recent studies have demonstrated the potential of LLMs in interpreting biological data by treating gene symbols as natural language. For instance, models like GPT-4 can perform zero-shot cell type annotation with remarkable accuracy by simply inputting lists of marker genes \cite{hou2024assessing, zeng2023revolutionizing}. In addition, LLMs are capable of generating coherent biological summaries and GO \cite{ashburner2000gene} terms, often rivaling the performance of human experts \cite{huang2024qust, hu2025evaluation}. From the perspective of semantic embeddings, GenePT \cite{chen2025simple} demonstrates the capability of GPT models to generate informative embeddings from either gene description texts or gene symbols alone. When integrating transcriptomic data with histopathological images, SGN \cite{yang2024spatial} performs dot-product operations between LLM-derived gene embeddings and visual features, enabling zero-shot gene expression prediction. Similarly, OmiCLIP \cite{chen2025visual} employs a contrastive learning framework to align gene semantics with image patch features within a unified feature space, thereby establishing a visual-omics foundation model. These methods all successfully leverage the language processing power of LLMs to unlock biological insights, demonstrating the potential of incorporating linguistic information into spatial transcriptomic data analysis.

\subsection{Spatial Domain Identification}
The primary goal of spatial domain identification in spatial transcriptomics is to cluster spots with similar characteristics, a task crucial for understanding tissue architecture. Early approaches such as Louvain \cite{blondel2008fast} and Seurat \cite{satija2015spatial} typically overlooked spatial context. To address this, methods specifically designed to incorporate spatial information have emerged. BayesSpace \cite{zhao2021spatial} employs a Bayesian model with a spatial prior to encourage neighboring spots to belong to the same cluster. SpaGCN \cite{hu2021spagcn} was a pioneering deep learning approach, constructing a graph from the tissue and using a Graph Convolutional Network (GCN) to integrate gene expression with spatial coordinates. Building on this, a family of GNN-based models has become a dominant paradigm. STAGATE \cite{dong2022deciphering} uses a graph attention autoencoder to learn low-dimensional embeddings that capture both expression similarity and spatial proximity. GraphST \cite{long2023spatially} further enhances this by incorporating contrastive learning to better distinguish between different spatial domains. Spatial-MGCN \cite{wang2023spatial} employs a multi-view GCN to model the complex relationships between gene expression and spatial information. However, a limitation persists across all these methods: they treat gene expression as a numerical matrix, discarding the rich biological knowledge encoded in gene symbols.

\section{Method}

\begin{figure*}[t]
\centering
\includegraphics[width=0.95\textwidth]{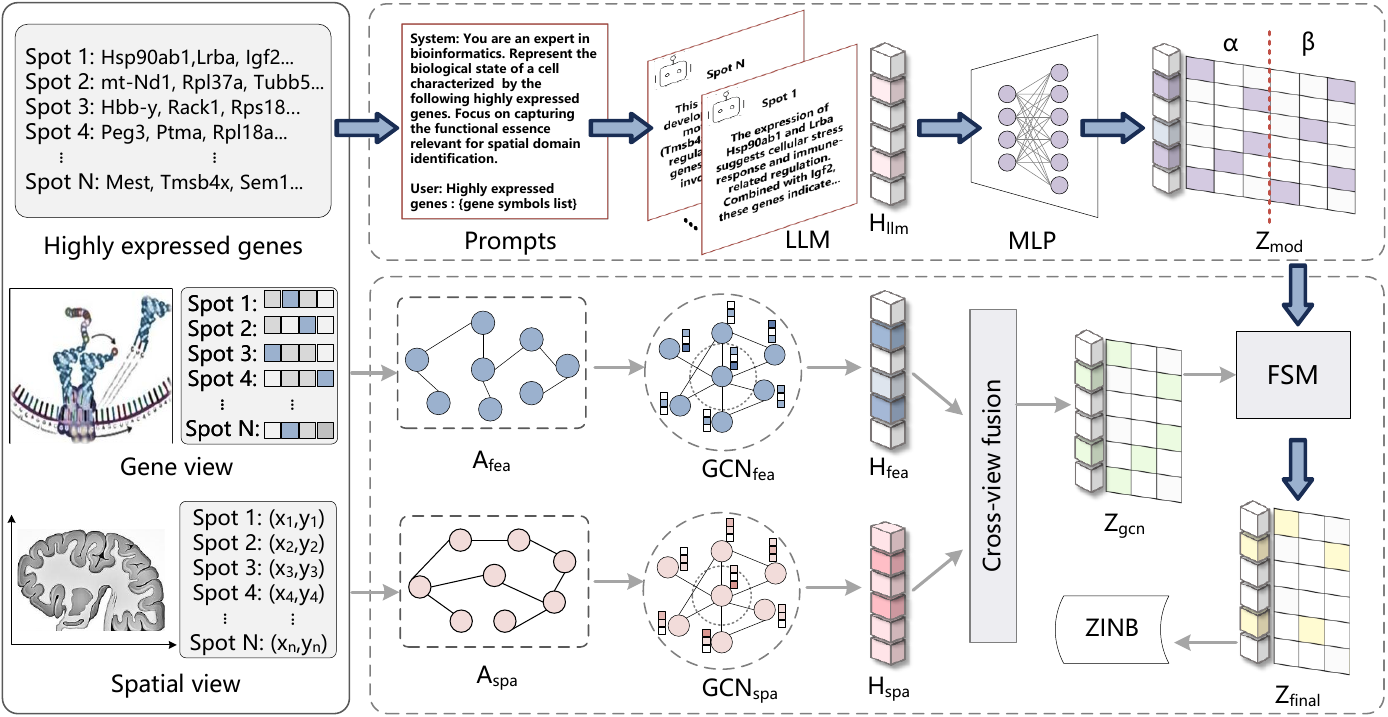}
\caption{The framework of the proposed SemST. We integrate three distinct information as inputs. Two graphs are constructed based on expression and spatial data, and multi-view features are propagated and fused using GCN. Gene symbols are formatted into prompts and fed into an LLM to extract biologically informed embeddings, which are then used by the proposed FSM module to guide spatial feature refinement. Model optimization is achieved through data reconstruction based on ZINB.}
\label{fig:framework}
\end{figure*}

In this section, we present SemST, a novel semantic-guided deep learning framework for spatial transcriptomics data clustering. The overall architecture of SemST is illustrated in Figure~\ref{fig:framework}. 

\subsection{Problem Formulation}
We consider a spatial transcriptomics dataset denoted as $\mathcal{D} = \{\mathbf{X}, \mathbf{S}, \mathcal{G}\}$. The normalized gene expression matrix is represented by $\mathbf{X} \in \mathbb{R}^{N \times M}$, where $N$ is the number of spots and $M$ is the number of genes. The spatial coordinates of the spots are given by $\mathbf{S} \in \mathbb{R}^{N \times 2}$. The set of gene symbols is denoted by $\mathcal{G} = \{g_1, g_2, \dots, g_M\}$. The primary objective of our framework is to learn a low-dimensional feature representation for each spot, such that these representations can be effectively utilized for clustering to reveal underlying spatial domains and functional regions within the tissue.

\subsection{Multi-View GNN for Spatial Representation}
To comprehensively capture the relationships among spots, we construct two distinct graphs: one based on spatial proximity and the other on gene expression similarity. The former excels at modeling microenvironmental interactions, while the latter is well suited for identifying spots that belong to the same layer but are spatially distant.

\paragraph{Spatial Proximity Graph.}
We construct a spatial graph $\mathbf{A}_{spa} \in \{0, 1\}^{N \times N}$ based on the physical coordinates $\mathbf{S}$. An edge is established between two spots $i$ and $j$ if their Euclidean distance is within a predefined radius threshold $r$. The adjacency matrix is formally defined as:
\begin{equation}
({\textbf{A}}_{spa})_{ij} = 
\begin{cases} 
1, & \text{if } \|\mathbf{S}_i - \mathbf{S}_j\|_2 \leq r, \\
0, & \text{otherwise}.
\end{cases}
\end{equation}

\paragraph{Gene Expression Similarity Graph.}
We construct a feature graph $\mathbf{A}_{fea} \in \{0, 1\}^{N \times N}$ using the K-Nearest Neighbors (KNN) \cite{cover1967nearest} algorithm on the gene expression matrix $\mathbf{X}$. An edge connects spot $i$ to spot $j$ if spot $j$ is among the top $k_{n}$ most similar spots to spot $i$ in the high-dimensional expression space. This is defined as:
\begin{equation}
(\textbf{A}_{fea})_{ij} = 
\begin{cases} 
1, & \text{if } j \in \mathcal{N}(i), \\
0, & \text{otherwise},
\end{cases}
\end{equation}
where $\mathcal{N}(i)$ represents the set of $k_{n}$ nearest neighbors of spot $i$ based on the cosine similarity of their expression profiles $\mathbf{X}_i$ and $\mathbf{X}_j$.

\paragraph{Multi-View GCN Propagation.}
With the two graphs constructed, we employ a multi-view GCN to learn latent representations from both spatial and feature perspectives. Each branch processes the gene expression matrix $\mathbf{X}$ as node features but uses a different graph structure.

The propagation rule for a GCN layer is given by:
\begin{equation}
\mathbf{H}^{(l+1)} = \sigma(\tilde{\mathbf{D}}^{-\frac{1}{2}}\tilde{\mathbf{A}}\tilde{\mathbf{D}}^{-\frac{1}{2}}\mathbf{H}^{(l)}\mathbf{W}^{(l)}),
\end{equation}
where $\tilde{\mathbf{A}} = \mathbf{A} + \mathbf{I}$ is the adjacency matrix with self-loops, $\tilde{\mathbf{D}}$ is the corresponding degree matrix, $\mathbf{H}^{(l)}$ is the feature matrix at layer $l$ ($\mathbf{H}^{(0)} = \mathbf{X}$), $\mathbf{W}^{(l)}$ is a trainable weight matrix at layer $l$, and $\sigma$ is a non-linear activation function.

The two branches generate embeddings as follows:
\begin{align}
\mathbf{H}_{spa} &= \text{GCN}_{spa}(\mathbf{X}, \mathbf{A}_{spa}), \\
\mathbf{H}_{fea} &= \text{GCN}_{fea}(\mathbf{X}, \mathbf{A}_{fea}),
\end{align}
where $\text{GCN}_{spa}$ and $\text{GCN}_{fea}$ are multi-layer GCNs with separate parameters. The fused spatial representation, $\mathbf{Z}_{gcn} \in \mathbb{R}^{N \times d}$, is obtained by concatenating the outputs of the two branches:
\begin{equation}
\mathbf{Z}_{gcn} = \text{Concat}(\mathbf{H}_{spa}, \mathbf{H}_{fea}).
\end{equation}

This multi-view approach allows the model to learn a robust representation that jointly captures local spatial proximity and distance-independent expression profile similarity.

\subsection{Semantic-Guided Feature Modulation}
Fundamental to SemST is its ability to leverage the rich biological knowledge inherent in gene symbols. We achieve this by generating semantic embeddings using a general LLM and then using these embeddings to modulate the GNN-derived representation $\mathbf{Z}_{gcn}$ through our FSM module.

\paragraph{Gene Semantics Extraction.}
For each spot $i$, we begin by identifying its key biological characteristics through the selection of the top $k_{g}$ most highly expressed genes, denoted as $\mathcal{G}_i^k = \{g_{i,1}, g_{i,2}, \dots, g_{i,k_{g}}\}$. These gene symbols are formatted into a natural language string and incorporated into a descriptive prompt, which is subsequently fed into an LLM. The parameters of LLM remain frozen throughout training, allowing it to serve as a fixed, comprehensive biological knowledge base. From the LLM, we extract the final-layer hidden state, denoted as $\mathbf{H}_{llm} \in\mathbb{R}^{N \times d'}$, which encodes rich, high-order biological knowledge.

\paragraph{Fine-grained Semantic Modulation (FSM).}

Integrating semantic and spatial information is crucial for comprehensive biological understanding. While conventional fusion strategies (e.g., concatenation, addition, cross-attention) are commonly employed, they often exhibit limited efficacy in fully leveraging the biological prior knowledge from the LLM.
A more fine-grained and expressive modulation is required to facilitate semantic-spatial integration \cite{perez2018film}. The proposed FSM module specifically aims to adapt the powerful LLM-derived biological prior for our task.
Firstly, we project the $\mathbf{H}_{llm}$ through a trainable Multi-Layer Perceptron (MLP). This MLP transforms the high-dimensional $\mathbf{H}_{llm}$ into a task-specific modulation space, yielding the semantic modulation matrix $\mathbf{Z}_{mod} \in \mathbb{R}^{N \times 2d}$ with an output dimension of $2 \times d$:
\begin{equation}
\mathbf{Z}_{mod} = f_{mlp}(\mathbf{H}_{llm}).
\end{equation}

Subsequently, splitting $\mathbf{Z}_{mod}$ along its feature dimension into two equal-sized tensors: a scaling factor $\bm{\alpha} \in \mathbb{R}^{N \times d}$ and a bias factor $\bm{\beta} \in \mathbb{R}^{N \times d}$:
\begin{equation}
[ \bm{\alpha} \mid \bm{\beta} ] = \mathbf{Z}_{{mod}}. \\
\end{equation}

% The split operation on unified representation ${\mathbf{Z}_{mod}}$ ensures that both $\bm{\alpha}$ and $\bm{\beta}$ inherently share the same semantic projection space, thereby promoting more coordinated and consistent spatial feature modulation. These factors then modulate the spatial embedding $\mathbf{Z}_{gcn}$ via an affine transformation with a residual connection. The final, semantically-calibrated representation $\mathbf{Z}_{final}$ is computed as:
% \begin{equation}
% \mathbf{Z}_{final} = ((1+\bm{\alpha}) \odot \mathbf{Z}_{gcn} + \bm{\beta}) \mathbf{W},
% \end{equation}
% where $\odot$ denotes the element-wise (Hadamard) product, and $\mathbf{W}$ is a trainable weight matrix. This affine transformation enables the semantic representation to dynamically modulate each spot’s spatial features through fine-grained scaling and shifting, effectively leveraging the biological knowledge captured by the LLM to actively steer the representation—letting the genes “speak” in the learning process.

The final, semantically-calibrated representation $\mathbf{Z}_{final}$ is computed as:
\begin{equation}
\mathbf{Z}_{final} = ((1+\bm{\alpha}) \odot \mathbf{Z}_{gcn} + \bm{\beta}) \mathbf{W},
\end{equation}
where $\odot$ denotes the element-wise (Hadamard) product, and $\mathbf{W}$ is a trainable weight matrix. This affine transformation enables the semantic representation to dynamically modulate each spot’s spatial features through fine-grained scaling and shifting, effectively leveraging the biological knowledge captured by the LLM to actively steer the representation—letting the genes “speak” in the learning process. 

Notably, we intentionally generate $\bm{\alpha}$ and $\bm{\beta}$ from the same MLP to ensure they are embedded in the same bio-semantic space. During backpropagation, the distinct gradients for scaling and shifting lead them to gradually differentiate and specialize, effectively decoupling identical semantics into distinct functions. Using two separate networks to generate $\bm{\alpha}$ and $\bm{\beta}$ risks uncoordinated modulation.

\subsection{Model Optimization}
To train the entire framework without ground-truth labels, we adopt a self-supervised learning strategy based on data reconstruction. 

\paragraph{ZINB Loss.}
Given the sparse and over-dispersed nature of spatial transcriptomics data, we use a Zero-Inflated Negative Binomial (ZINB) reconstruction loss \cite{yu2022zinb}. The decoder network transforms the final embedding $\mathbf{Z}_{{final}}$ into the three parameter matrices of the ZINB distribution for each gene in each spot: the mean $\bm{\mu} = f_{\mu}(\mathbf{Z}_{{final}})$, the dispersion $\bm{\theta} = f_{\theta}(\mathbf{Z}_{{final}})$, and the dropout probability $\bm{\pi} = f_{\pi}(\mathbf{Z}_{{final}})$. The ZINB loss is the negative log-likelihood of the observed expression counts $\mathbf{X}$ given these parameters:
\begin{equation}
\mathcal{L}_{{zinb}} = -\sum_{i=1}^{N}\sum_{j=1}^{M} \log P(\textbf{X}_{ij} \mid \bm{\mu}_{ij}, \bm{\theta}_{ij}, \bm{\pi}_{ij}).
\end{equation}

Furthermore, we adopt two additional loss functions, each of which has been demonstrated to be highly effective in prior studies \cite{zhu2024multi, he2024heterogeneous}.

\paragraph{Correlation Reduction Loss.} 

\begin{figure*}[t]
\centering
\includegraphics[width=0.95\textwidth]{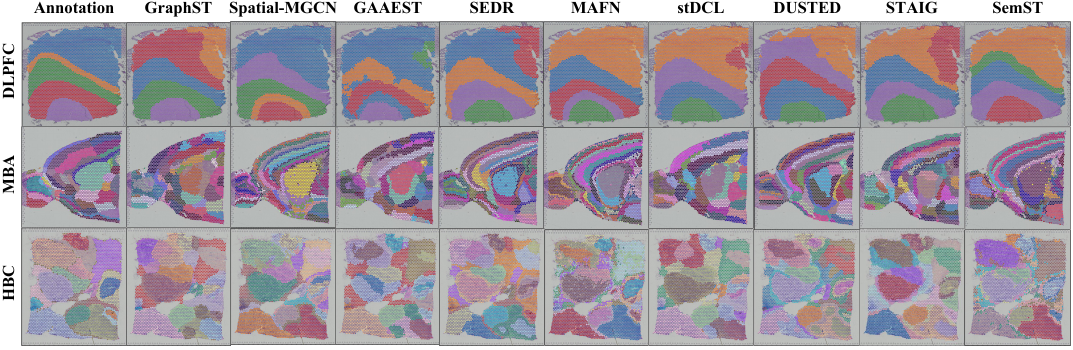}
\caption{Visualization of manual annotations and clustering results produced by SemST and eight other methods on the DLPFC slice \#151672, MBA, and HBC datasets. Color indicates spatial domains.}
\label{fig2}
\end{figure*}

\begin{table*}[t]
\centering
\small
\setlength{\tabcolsep}{1mm}

\begin{tabular}{c|cccc|cccc|cccc|cccc}
\toprule
{Datasets} & \multicolumn{4}{c|}{{151508}} & \multicolumn{4}{c|}{{151509}} & \multicolumn{4}{c|}{{151510}} & \multicolumn{4}{c}{{151671}} \\
\midrule
{Methods} & ARI & NMI & ACC & F1 & ARI & NMI & ACC & F1 & ARI & NMI & ACC & F1 & ARI & NMI & ACC & F1 \\
\midrule
\text{STAGATE}{\scriptsize\textcolor{gray}{[Nat. Com.'22]}} & 53.84 & 64.10 & 69.39 & \underline{64.13} & 49.54 & 65.81 & 67.42 & 64.53 & 46.07 & 59.17 & 61.55 & 50.88 & 59.54 & 69.37 & 75.10 & 77.56 \\
\text{GraphST}{\scriptsize\textcolor{gray}{[Nat. Com.'23]}} & 48.61 & 64.28 & 61.93 & 56.78 & 51.93 & 67.41 & 67.48 & 63.40 & 51.29 & 64.53 & 68.38 & 53.35 & 61.02 & 71.77 & 74.84 & 79.62 \\
\text{Spatial-MGCN}{\scriptsize\textcolor{gray}{(BIB'23)}} & 46.22 & 60.23 & 59.46 & 56.45 & 54.22 & 68.16 & 63.83 & 63.27 & 51.61 & 66.50 & 59.35 & 52.38 & 60.19 & 71.87 & 73.39 & 75.18 \\
\text{GAAEST}{\scriptsize\textcolor{gray}{[Com. Biol.'24]}} & 31.24 & 52.82 & 57.18 & 44.79 & 43.84 & 53.16 & 67.21 & 43.39 & 39.79 & 53.21 & 59.61 & 45.30 & 64.80 & 65.89 & 74.47 & 71.41 \\ 
\text{SEDR}{\scriptsize\textcolor{gray}{[Genome Med.'24]}} & 47.47 & 62.07 & 61.79 & 60.53 & 49.69 & 63.30 & 66.54 & 58.36 & 50.52 & 64.76 & 60.20 & 60.70 & 59.97 & 68.25 & 74.44 & 76.96 \\
\text{MAFN}{\scriptsize\textcolor{gray}{[TKDE'24]}} & 54.54 & 66.31 & 66.26 & 62.61 & \underline{70.48} & 69.41 & \underline{75.36} & 66.07 & \underline{69.91} & \underline{70.67} & \underline{74.82} & 64.07 & \underline{72.41} & 68.05 & \underline{79.53} & {79.29} \\
\text{stDCL}{\scriptsize\textcolor{gray}{[Adv. Sci.'25]}} & \underline{57.13} & \underline{66.78} & \underline{71.42} & 60.48 & 41.63 & 61.41 & 62.87 & 53.99 & {57.91} & 62.49 & {71.95} & \textbf{71.95} & {66.74} & \underline{73.91} & {79.23} & \underline{81.30} \\
\text{DUSTED}{\scriptsize\textcolor{gray}{[AAAI'25]}} & 46.53 & 59.12 & 64.00 & {61.63} & 51.10 & 66.38 & 67.08 & 63.68 & 43.08 & 62.76 & 64.35 & 58.60 & 60.31 & 69.29 & 75.76 & {75.94} \\
\text{STAIG}{\scriptsize\textcolor{gray}{[Nat. Com.'25]}} & 50.36 & 64.09 & 63.02 & 57.19 & {59.45} & \underline{69.51} & {71.60} & \underline{67.42} & 52.88 & {67.02} & 65.46 & 60.33 & 45.17 & 65.56 & 59.25 & 58.48 \\
\midrule
\text{SemST} & \textbf{67.93} & \textbf{71.08} & \textbf{74.53} & \textbf{67.59} & \textbf{73.90} & \textbf{72.73} & \textbf{76.57} & \textbf{67.49} & \textbf{73.92} & \textbf{73.09} & \textbf{77.30} & \underline{68.29} & \textbf{78.70} & \textbf{75.20} & \textbf{81.21} & \textbf{82.01} \\
\midrule
\midrule
{Datasets} & \multicolumn{4}{c|}{{ME}} & \multicolumn{4}{c|}{{MVC}} & \multicolumn{4}{c|}{{HBC}} & \multicolumn{4}{c}{{MBA}} \\
\midrule
{Methods} & ARI & NMI & ACC & F1 & ARI & NMI & ACC & F1 & ARI & NMI & ACC & F1 & ARI & NMI & ACC & F1 \\
\midrule
\text{STAGATE}{\scriptsize\textcolor{gray}{[Nat. Com.'22]}} & 32.34 & 55.51 & 46.07 & 47.77 & 52.38 & 56.37 & 66.86 & 65.55 & 44.66 & 67.36 & 48.97 & 53.50 & 35.81 & \textbf{72.49} & 45.49 & 48.80 \\
\text{GraphST}{\scriptsize\textcolor{gray}{[Nat. Com.'23]}} & 29.75 & 53.54 & 46.12 & 44.95 & 36.16 & 45.61 & 60.73 & 61.69 & 52.63 & 66.96 & \underline{64.98} & 55.58 & 41.32 & \underline{71.46} & 47.05 & 48.23 \\
\text{Spatial-MGCN}{\scriptsize\textcolor{gray}{(BIB'23)}} & \underline{44.54} & \underline{59.91} & \underline{56.79} & \underline{57.38} & {53.32} & {59.89} & {74.31} & 74.40 & \underline{65.68} & \textbf{70.83} & 64.67 & \underline{65.71} & \underline{48.34} & 68.03 & 44.12 & 44.87 \\
\text{GAAEST}{\scriptsize\textcolor{gray}{[Com. Biol.'24]}} & 26.27 & 49.18 & 44.78 & 45.45 & \underline{59.41} & {68.15} & 75.89 & 73.75 & 52.02 & 67.22 & 55.71 & 55.97 & 43.35 & 70.64 & \underline{50.09} & 49.29 \\
\text{SEDR}{\scriptsize\textcolor{gray}{[Genome Med.'24]}} & 27.70 & 51.32 & 46.32 & 46.89 & 52.71 & 63.16 & 67.77 & 66.17 & 43.16 & 67.30 & 50.34 & 54.19 & 40.36 & 71.37 & 46.86 & \underline{50.19} \\
\text{MAFN}{\scriptsize\textcolor{gray}{[TKDE'24]}} & 38.48 & 54.83 & {53.37} & {54.37} & {59.12} & {67.37} & \underline{76.97} & \underline{74.91} & 59.20 & 62.26 & 60.61 & 61.34 & 44.15 & 67.73 & 44.64 & 45.48 \\
\text{stDCL}{\scriptsize\textcolor{gray}{[Adv. Sci.'25]}} & {34.32} & {59.34} & {51.68} & {50.69} & 49.54 & 63.49 & {66.94} & 62.17 & 55.73 & 67.80 & 58.19 & 56.21 & 42.05 & 70.75 & 48.24 & 45.06 \\
\text{DUSTED}{\scriptsize\textcolor{gray}{[AAAI'25]}} & 26.23 & 52.15 & 40.03 & {39.67} & 58.06 & 66.30 & 71.25 & {71.33} & 47.81 & 65.78 & 48.92 & 52.40 & 35.86 & 71.13 & 43.75 & 47.31 \\
\text{STAIG}{\scriptsize\textcolor{gray}{[Nat. Com.'25]}} & 27.20 & 51.02 & 46.68 & 47.62 & 58.26 & \underline{69.58} & 71.91 & 70.61 & 57.86 & \underline{69.43} & 59.82 & 63.19 & 33.35 & 70.61 & 44.12 & 46.68 \\
\midrule
\text{SemST} & \textbf{50.50} & \textbf{60.06} & \textbf{60.93} & \textbf{59.07} & \textbf{64.27} & \textbf{70.60} & \textbf{80.70} & \textbf{79.91} & \textbf{68.64} & 68.23 & \textbf{67.27} & \textbf{65.99} & \textbf{53.92} & {70.12} & \textbf{51.09} & \textbf{50.60} \\
\bottomrule
\end{tabular}
\caption{Quantitative comparison of clustering performance on eight spatial transcriptomics datasets using four evaluation metrics: ARI, NMI, ACC, and F1-score. \textbf{Bold} denotes the best result, and \underline{Underline} denotes the second best.}
\label{tab:clustering_all}
\end{table*}

To strengthen the GNN backbone, we introduce correlation reduction loss. Specifically, given two embeddings ${\mathbf{H}}_{spa}$ and ${\mathbf{H}}_{fea}$, we compute the cosine similarity between the $i$-th feature in spatial view and the $j$-th feature in the feature view:
\begin{equation}
\textbf{C}_{ij} = \frac{({\mathbf{H}}^{spa}_{i})({\mathbf{H}}^{fea}_j)^\top}{\|{\mathbf{H}}^{spa}_i\| \, \|{\mathbf{H}}^{fea}_j\|}.
\end{equation}
Then, the correlation reduction loss encourages correlation matrix $\mathbf{C} \in \mathbb{R}^{p \times p}$ (where $p = d/2$) to be close to the identity matrix $\mathbf{I}$, thereby promoting consistence of corresponding features across views while reducing correlations between distinct features. The loss is formulated as:
\begin{equation}
\mathcal{L}_{cr} = \frac{1}{p^2} \sum_{i,j} (\textbf{C}_{ij} - \textbf{I}_{ij})^2.
\end{equation}

\paragraph{Spatial Regularization Loss.}
% To ensure that spatially neighboring spots are close and non-neighboring spots are distant in the latent space, we introduce a spatial regularization loss that integrates both similarity and spatial adjacency:

We introduce a spatial regularization loss to prevent the over-injection of semantics that may harm spatial coherence. This loss ensures that spatially neighboring spots are close, while non-neighboring spots are distant in the latent space $\mathbf{Z}_{final}$, formulated as:

\begin{equation}
\mathcal{L}_{{s}} = - \sum_{i=1}^{N} \left( 
\sum_{j \in \mathcal{R}_i} \log \sigma(\psi_{ij}) + 
\sum_{k \notin \mathcal{R}_i} \log \big(1 - \sigma(\psi_{ik})\big) 
\right),
\end{equation}
where $\psi_{ij}$ denotes the cosine similarity between the representations of spots $i$ and $j$, $\sigma(\cdot)$ denotes the sigmoid function and $\mathcal{R}_i$ represents the set of spatial neighbors of spot $i$.

\paragraph{Overall Loss Function.}
The total loss function for SemST is formulated as:
\begin{equation}
\mathcal{L} = \mathcal{L}_{zinb} + \gamma\mathcal{L}_{cr} + \lambda\mathcal{L}_{s},
\end{equation}
where $\gamma$ and $\lambda$ are hyper-parameters that balance the contributions of the different loss terms.

\section{Experiments}

\subsection{Experimental Setup}
\paragraph{Datasets.}
We evaluate our method on nine public spatial transcriptomics datasets. The human dorsolateral prefrontal cortex dataset (DLPFC), the human breast cancer dataset (HBC), and the mouse brain anterior dataset (MBA) are all generated using the 10x Genomics Visium platform. For the DLPFC dataset, we use five tissue slices—151508, 151509, 151510, 151671, and 151672—obtained from two adult human samples, with each section manually annotated into 5 or 7 cortical layers. The HBC dataset is annotated into 20 spatial domains, and the MBA dataset contains 52 annotated spatial domains. We also include the mouse embryo dataset (ME) from the Stereo-seq platform, which is derived from an E9.5 sample and includes 12 annotated spatial domains, and the mouse visual cortex dataset (MVC) from the STARmap platform, which contains 7 annotated spatial domains.

\paragraph{Compared Methods.}
We compare our method with nine representative methods, including STAGATE \cite{dong2022deciphering}, GraphST \cite{long2023spatially}, Spatial-MGCN \cite{wang2023spatial}, GAAEST \cite{wang2024graph}, SEDR \cite{xu2024unsupervised}, MAFN \cite{zhu2024multi}, stDCL \cite{yu2025accurate}, DUSTED \cite{zhu2025dusted}, and STAIG \cite{yang2025staig}. All baseline results are obtained by running official implementations with recommended parameters to achieve the best performance.

\paragraph{Metrics.}
We adopt four standard clustering metrics to evaluate performance: Adjusted Rand index (ARI), Normalized Mutual Information (NMI), clustering accuracy (ACC), and F1-score (F1). These metrics comprehensively assess the agreement between the predicted clusters and ground-truth annotations. We multiplied all the results by 100.

\paragraph{Implementation Details.}
Our model is implemented in PyTorch and trained using the Adam optimizer with an initial learning rate of $0.001$ and a weight decay of $5 \times 10^{-4}$.
% Qwen3-4B \cite{yang2025qwen3} is used as the LLM to provide biological prior knowledge. The final clustering is performed using the K-means algorithm. 
All experiments are conducted on a machine equipped with an NVIDIA RTX 3090 GPU (24GB). More implementation details can be found in the supplementary materials.

\begin{table*}[!htbp]
\centering
\label{tab:ablation-transposed}
\small % To help fit the wide table
\setlength{\tabcolsep}{1.0mm} % Reduce column separation to help fit
\begin{tabular}{c|cccc|cccc|cccc|cccc}
\toprule

\multirow{2}{*}{Variants} & \multicolumn{4}{c|}{151509} & \multicolumn{4}{c|}{151510} & \multicolumn{4}{c|}{151671} & \multicolumn{4}{c}{MBA} \\  \cmidrule(l){2-17}
& ARI    & NMI    & ACC   & F1  & ARI    & NMI    & ACC   & F1 & ARI    & NMI    & ACC   & F1 & ARI    & NMI    & ACC   & F1 \\
\midrule
w/o LLM  & 67.37 & 69.16 & 74.02 & 67.13 & 70.74 & 71.40 & 75.26 & 66.86 & 60.43 & 71.18 & 75.20 & 78.20 & 44.83 & 67.74 & 47.27 & 48.10 \\
\midrule
Random Emb. & 59.95 & 59.45 & 69.61 & 60.74 & 63.16 & 60.84 & 73.23 & 54.40 & 50.56 & 62.79 & 67.99 & 71.35 & 41.72 & 65.35 & 42.97 & 41.65 \\
Unrelated Emb. & 65.24 & 68.23 & 71.60 & 61.85 & 65.96 & 68.50 & 74.67 & 66.39 & 60.11 & 72.62 & 72.67 & 76.71 & 43.89 & 68.20 & 46.79 & 48.63 \\
BERT & 63.27 & 65.71 & 68.34 & 60.19 & 69.25 & 70.30 & 75.50 & 67.28 & 60.05 & 70.67 & 73.93 & 77.40 & 42.96 & 68.35 & 45.86 & 46.61 \\
\midrule
Cross-Attention     & 62.40 & 56.75 & 64.68 & 66.26 & 66.07 & 67.48 & 72.36 & 64.70 & 63.14 & 74.03 & 75.64 & 78.87 & 47.69 & 67.45 & 48.50 & 49.25 \\
Concat    & 69.25 & 68.91 & 72.99 & 62.50 & 67.43 & 69.88 & 73.69 & 65.79 & 73.21 & 67.64 & 80.23 & 80.12 & 46.64 & 68.05 & 47.01 & 47.97 \\
Add     & 69.19 & 68.24 & 73.02 & 64.22 & 71.96 & 71.37 & 76.21 & 67.92 & 74.14 & 71.83 & 79.40 & 80.60 & 47.35 & 67.96 & 47.24 & 47.70 \\
\midrule
SemST     & \textbf{73.90} & \textbf{72.73} & \textbf{76.57} & \textbf{67.49} & \textbf{73.92} & \textbf{73.09} & \textbf{77.30} & \textbf{68.29} & \textbf{78.70} & \textbf{75.20} & \textbf{81.21} & \textbf{82.01} & \textbf{53.92} & \textbf{70.12} & \textbf{51.09} & \textbf{50.60} \\
\bottomrule
\end{tabular}
\caption{Ablation study evaluating the contributions of semantic embeddings and the FSM module.}
\label{tab:ablation}
\end{table*}

\sisetup{
    detect-all,
    table-align-text-post = false,
    table-format=2.2,
    separate-uncertainty = true
}

\begin{table*}[!htbp]
\centering
\small
\setlength{\tabcolsep}{1mm}

\begin{tabular}{
    c| % Datasets 列
    l|
    S[table-format=2.2]@{\,/\,}S[table-format=2.2]@{\,/\,}S[table-format=2.2]
    S[table-format=2.2]@{\,/\,}S[table-format=2.2]@{\,/\,}S[table-format=2.2]
    S[table-format=2.2]@{\,/\,}S[table-format=2.2]@{\,/\,}S[table-format=2.2]
    S[table-format=2.2]@{\,/\,}S[table-format=2.2]@{\,/\,}S[table-format=2.2]
}

% \begin{tabular}{
%     c| % Datasets 列
%     l|
%     ccc|
%     ccc|
%     ccc|
%     ccc
% }

\toprule
\multirow{2}{*}{Datasets} & \multirow{2}{*}{Methods} 
& \multicolumn{3}{c}{ARI} & \multicolumn{3}{c}{NMI} & \multicolumn{3}{c}{ACC} & \multicolumn{3}{c}{F1} \\
\cmidrule(lr){3-5} \cmidrule(lr){6-8} \cmidrule(lr){9-11} \cmidrule(lr){12-14} & 
& {Before} & {After} & {$\Delta$}
& {Before} & {After} & {$\Delta$}
& {Before} & {After} & {$\Delta$}
& {Before} & {After} & {$\Delta$} \\
\midrule
\multirow{9}{*}{HBC} 
& STAGATE {\scriptsize\textcolor{gray}{[Nat. Com.'22]}} & 44.66 & 58.16 & \textbf{+13.50} & 67.36 & 68.80 & \textbf{+1.44} & 48.97 & 58.77 & \textbf{+9.80} & 53.50 & 61.17 & \textbf{+7.67} \\
& GraphST {\scriptsize\textcolor{gray}{[Nat. Com.'23]}} & 52.63 & 57.09 & \textbf{+4.46} & 66.96 & 68.20 & \textbf{+1.24} & 54.98 & 58.79 & \textbf{+3.81} & 55.58 & 59.06 & \textbf{+3.48} \\
& Spatial-MGCN {\scriptsize\textcolor{gray}{[BIB'23]}} & 65.68 & 66.80 & \textbf{+1.12} & 70.83 & 69.44 & -1.39 & 64.67 & 65.61 & \textbf{+0.94} & 65.71 & 66.61 & \textbf{+0.90} \\
& GAAEST {\scriptsize\textcolor{gray}{[Com. Biol.'24]}} & 52.02 & 57.75 & \textbf{+5.73} & 67.22 & 68.29 & \textbf{+1.07} & 55.71 & 59.87 & \textbf{+4.16} & 55.97 & 58.95 & \textbf{+2.98} \\
& SEDR {\scriptsize\textcolor{gray}{[Genome Med.'24]}} & 43.16 & 50.17 & \textbf{+7.01} & 67.30 & 68.57 & \textbf{+1.27} & 50.34 & 56.27 & \textbf{+5.93} & 54.19 & 60.30 & \textbf{+6.11} \\
& MAFN {\scriptsize\textcolor{gray}{[TKDE'24]}} & 57.49 & 62.68 & \textbf{+5.19} & 62.78 & 64.58 & \textbf{+1.80} & 58.32 & 62.74 & \textbf{+4.42} & 59.55 & 62.47 & \textbf{+2.92} \\
& stDCL {\scriptsize\textcolor{gray}{[Adv. Sci.'25]}} & 55.73 & 62.10 & \textbf{+6.37} & 70.05 & 70.22 & \textbf{+0.17} & 58.19 & 63.22 & \textbf{+5.03} & 56.21 & 61.62 & \textbf{+5.41} \\
& DUSTED {\scriptsize\textcolor{gray}{[AAAI'25]}} & 47.81 & 50.95 & \textbf{+3.14} & 65.78 & 66.20 & \textbf{+0.42} & 48.92 & 52.24 & \textbf{+3.32} & 52.40 & 56.24 & \textbf{+3.84} \\
& STAIG {\scriptsize\textcolor{gray}{[Nat. Com.'25]}} & 57.86 & 60.33 & \textbf{+2.47} & 69.43 & 70.59 & \textbf{+1.16} & 59.82 & 61.11 & \textbf{+1.29} & 63.19 & 64.32 & \textbf{+1.13} \\
\midrule
\multirow{9}{*}{MBA} 
& STAGATE {\scriptsize\textcolor{gray}{[Nat. Com.'22]}} & 35.81 & 38.79 & \textbf{+2.98} & 72.49 & 72.11 & -0.38 & 45.49 & 46.64 & \textbf{+1.15} & 48.80 & 50.05 & \textbf{+1.25} \\
& GraphST {\scriptsize\textcolor{gray}{[Nat. Com.'23]}} & 41.32 & 48.82 & \textbf{+7.50} & 71.46 & 71.98 & \textbf{+0.52} & 47.05 & 50.58 & \textbf{+3.53} & 48.23 & 48.33 & \textbf{+0.10} \\
& Spatial-MGCN {\scriptsize\textcolor{gray}{[BIB'23]}} & 48.34 & 51.07 & \textbf{+2.73} & 68.03 & 68.90 & \textbf{+0.87} & 44.12 & 48.65 & \textbf{+4.53} & 44.87 & 49.56 & \textbf{+4.69} \\
& GAAEST {\scriptsize\textcolor{gray}{[Com. Biol.'24]}} & 43.35 & 46.52 & \textbf{+3.17} & 70.64 & 71.78 & \textbf{+1.14} & 50.09 & 51.32 & \textbf{+1.23} & 49.29 & 49.77 & \textbf{+0.48} \\
& SEDR {\scriptsize\textcolor{gray}{[Genome Med.'24]}} & 42.05 & 44.95 & \textbf{+2.90} & 71.37 & 71.97 & \textbf{+0.60} & 45.71 & 48.00 & \textbf{+2.29} & 45.90 & 48.02 & \textbf{+2.12} \\
& MAFN {\scriptsize\textcolor{gray}{[TKDE'24]}} & 44.15 & 48.06 & \textbf{+3.91} & 67.73 & 67.83 & \textbf{+0.10} & 44.64 & 47.72 & \textbf{+3.08} & 45.48 & 47.88 & \textbf{+2.40} \\
& stDCL {\scriptsize\textcolor{gray}{[Adv. Sci.'25]}} & 42.05 & 48.46 & \textbf{+6.41} & 70.75 & 71.77 & \textbf{+1.02} & 48.24 & 51.61 & \textbf{+3.37} & 45.06 & 47.40 & \textbf{+2.34} \\
& DUSTED {\scriptsize\textcolor{gray}{[AAAI'25]}} & 35.86 & 38.30 & \textbf{+2.44} & 71.13 & 71.68 & \textbf{+0.55} & 43.75 & 45.68 & \textbf{+1.93} & 47.13 & 48.38 & \textbf{+1.25} \\
& STAIG {\scriptsize\textcolor{gray}{[Nat. Com.'25]}} & 33.45 & 36.33 & \textbf{+2.98} & 70.61 & 71.47 & \textbf{+0.86} & 44.12 & 46.23 & \textbf{+2.11} & 46.68 & 49.00 & \textbf{+2.32} \\
\bottomrule
\end{tabular}
\caption{Clustering performance comparison before and after integrating our proposed module into existing methods.}
\label{tab:ablation-inte}
\end{table*}

\subsection{Quantitative and Qualitative Results}
Table~\ref{tab:clustering_all} presents the quantitative clustering results on eight spatial transcriptomics datasets. SemST almost consistently outperforms all compared methods across most metrics and datasets. For example, on the 151508 slice, SemST surpasses the closest competitor by over 10 ARI points. On the MBA dataset, characterized by highly complex spatial structures that make distinguishing small domains particularly challenging, SemST achieves an ARI improvement of more than 5 points over the best compared method.

Complementing these quantitative findings, Figure \ref{fig2} provides a qualitative visualization of spatial clustering results on three representative datasets. The visualizations clearly show that SemST produces more anatomically coherent and biologically meaningful clusters, aligning closely with annotated ground truth spatial domains.
These results demonstrate that SemST effectively captures both functional and structural tissue patterns, achieving superior accuracy and biological consistency.

\subsection{Ablation Study}

To systematically evaluate the contribution of the key architectural components of SemST and validate the versatility of the FSM module, we conducted comprehensive ablation experiments. Table \ref{tab:ablation} presents the performance of SemST variants on four representative datasets. Table \ref{tab:ablation-inte} shows the clustering performance improvement achieved by integrating the FSM module into all compared methods.

\paragraph{Absence of Semantic Embeddings.}
The “w/o LLM” variant represents SemST trained without LLM-derived semantic embeddings. As shown, removing the semantic embeddings leads to a performance drop across all metrics and datasets. 
% For instance, on dataset 151671, ARI drops from 78.70\% (SemST) to 60.43\% (w/o LLM). 
This confirms that the biologically informed semantic embeddings play a critical role in enabling the superior performance of SemST.

\begin{figure}[t]
\centering
\includegraphics[width=0.95\columnwidth]{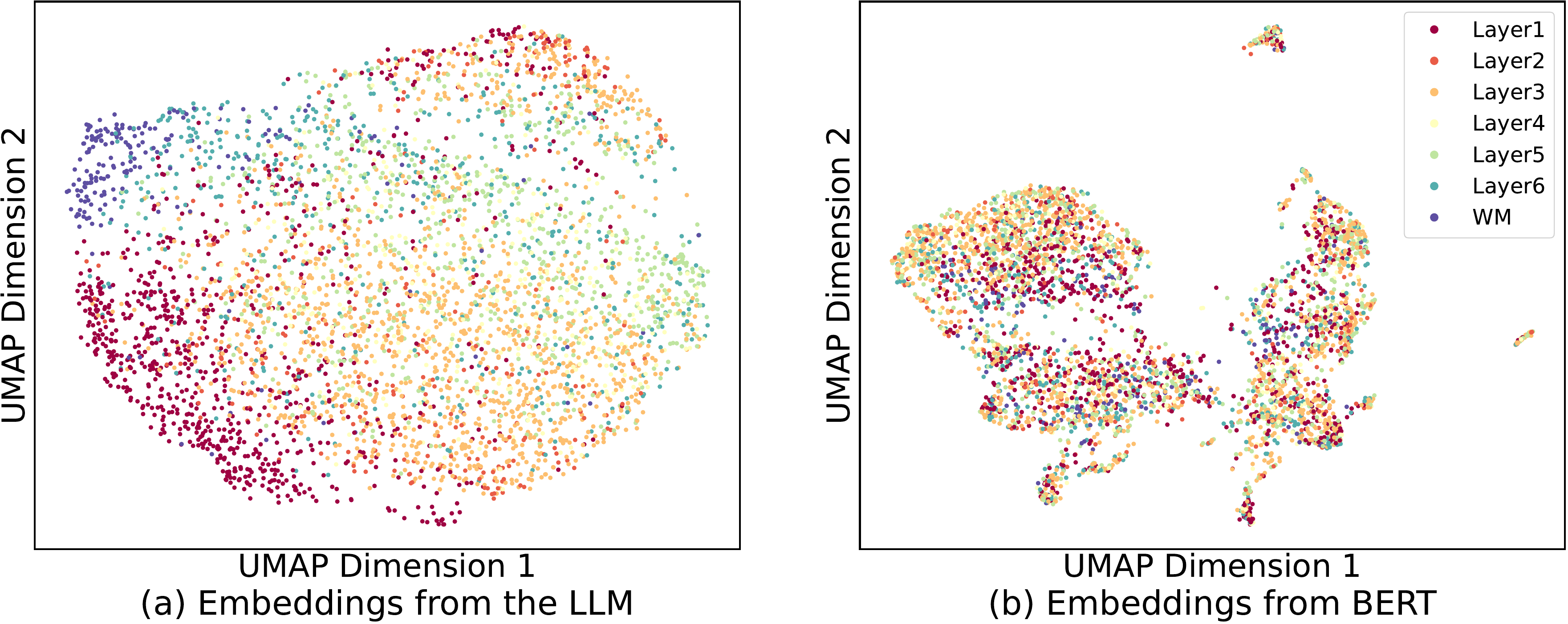}
\caption{UMAP visualization of semantic embeddings generated by the LLM and BERT on DLPFC slice \#151508 dataset, colored according to manual annotations.}
\label{umap}
\end{figure}

\paragraph{Quality of Semantic Embeddings.}
To further validate the importance of meaningful semantic information, we introduced three additional variants. “Random Emb.” involved randomly initialized vectors with a mean of 0 and a variance of 0.85, a distribution chosen to approximate the LLM embedding. For “Unrelated Emb.”, the LLM was prompted with entirely non-biological questions. The “BERT” variant merely employs a basic pre-trained BERT model \cite{devlin2019bert} as a plain text encoder, without incorporating any specialized domain knowledge. They almost consistently showed notably lower performance compared to the “w/o LLM” variant. This highlights the importance of using high-quality, biologically relevant semantic embeddings. Arbitrary or generic linguistic information may be ineffective or even detrimental.
The UMAP visualization (Figure \ref{umap}) demonstrates the superior quality of LLM-generated embeddings over that of BERT. The LLM captures a more smooth, biologically meaningful gradient across cortical layers (left), whereas the BERT embeddings are chaotic and lack clear structure (right). This illustrates that the LLM is capable of deriving deeper insights into gene symbols.

\paragraph{Effectiveness of FSM for Feature Fusion.}
We compared FSM module against several alternative fusion strategies for integrating semantic embeddings with spatial features, including “Cross-Attention”, “Concat” (concatenation), and “Add” (element-wise addition). SemST consistently outperforms all these alternatives, underscoring the superiority of FSM in enabling nuanced feature modulation. 
% Furthermore, most fusion strategies yield notable performance gains compared to non-fusion baselines, validating the effectiveness of incorporating gene semantic information in the spatial context.

\paragraph{Module Versatility.}
Based on the idea that biological semantics can universally enhance spatial representations, the FSM module is intended to function without reliance on any specific framework. To validate this plug-and-play versatility, we integrated semantic information via FSM into all baseline methods. As shown in Table \ref{tab:ablation-inte}, the results are compelling: the integration almost consistently yields significant performance improvements across nearly all methods and metrics, demonstrating the effectiveness of our approach in enriching spatial representations and improving performance on spatial domain identification.

\subsection{Parameters Analysis}

\begin{figure}[t]
\centering
\includegraphics[width=0.95\columnwidth]{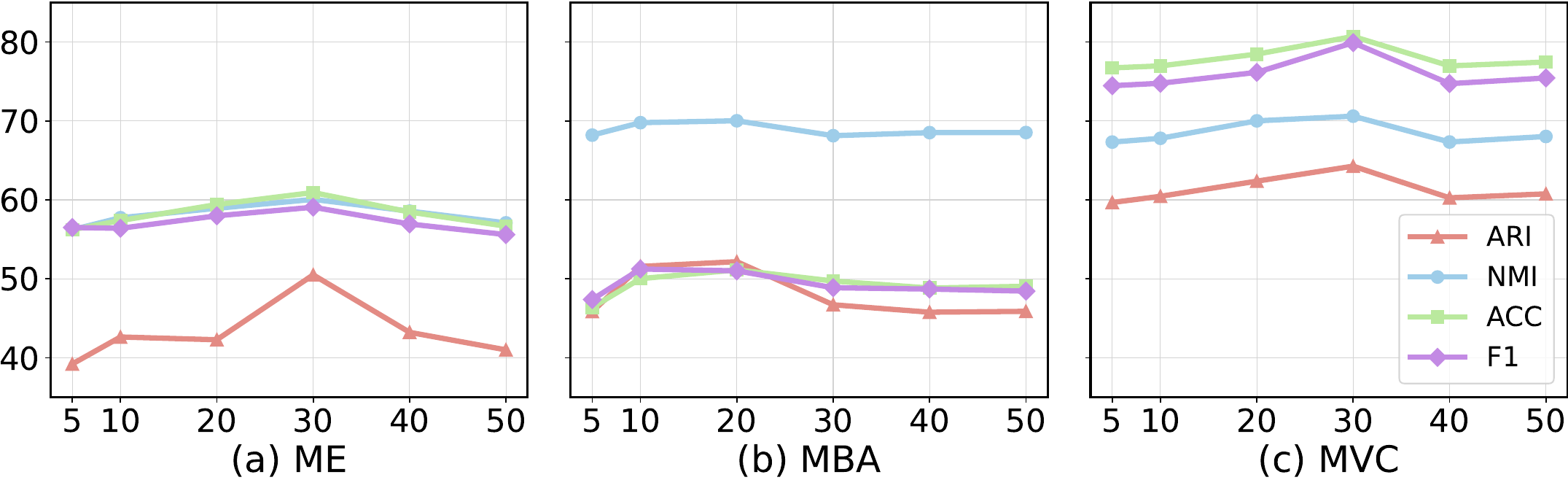}
\caption{Impact of $k_{g}$ (the number of top-expressed gene input to the LLM) on clustering performance across three datasets, evaluated by ARI, NMI, ACC, and F1.}
\label{curve}
\end{figure}

We examine how the number of top expressed genes ($k_g$) used as input affects clustering performance. As shown in Figure \ref{curve}, SemST was evaluated on three datasets with $k_g$ from 5 to 50. Performance generally improved with larger $k_g$, peaking at $k_g{=}30$ for ME and MVC, and $k_g{=}20$ for MBA, then declined. This suggests that a moderate-sized gene set best captures key biological signals, balancing information loss and gene redundancy.

\section{Conclusion}

In this work, we introduce SemST, a novel framework that addresses a critical limitation in spatial transcriptomics data clustering: the disregard for the biological semantics of gene symbols. By harnessing LLMs, SemST straightforwardly transforms the symbol sets of highly expressed genes into biologically rich semantic embeddings, which then function with the designed FSM module to modulate and guide spatial features. Remarkably, extensive experiments demonstrate that this simple yet effective approach leads to substantial performance improvements. By enabling genes to “speak” their biological roles within a spatial context, SemST offers a more profound and insightful perspective on the intricate tissue architecture.

\section{Acknowledgements}
The work was supported in part by the National Natural Science Foundation of China under grants 62476258 and 62522604, in part by the Natural Science Foundation of Hubei Province under grant 2025AFA113, and in part by the Fundamental Research Funds for National Universities, China University of Geosciences (Wuhan) under grant 2025XLB83.

\bibliography{SemST}

\clearpage

\twocolumn[
\begin{center}
\section*{Supplementary Material}
\end{center}
\vspace{3em} % 调整标题下间距
]

\section{Datasets and Preprocessing}

Table \ref{datasets} summarizes the datasets utilized throughout our experiments. To reduce technical noise, we begin by excluding spots located outside the main tissue regions in each spatial transcriptomics dataset. Next, we eliminate genes with low expression and low variability from the raw count matrices, retaining the top $g$ most variable genes. In particular, we set $g=128$ for the MVC dataset, while $g=3000$ is used for all other datasets. The resulting data is then normalized using a scaling factor as follows:
\begin{equation}
\mathbf{X}_{ij} = \frac{\text{count}_{ij}}{\sum_{j}\text{count}_{ij}} \times 10000.
\end{equation}

\section{More Implementation details}
For all operations in the experimental process that involve randomness, we set the random seed to 100 to ensure reproducible results. We apply the K-means clustering algorithm to the modulated latent representations to obtain the final spatial domain results.

There are five hyperparameters in our method: $r$ and $k_{n}$ for constructing the adjacency matrix, $k_{g}$ for generating semantic embedding, and $\gamma$ and $\lambda$ for balancing the loss items. We follow the hyperparameter settings used in previous works \cite{dong2022deciphering, wang2023spatial,zhu2024multi}. For the DLPFC dataset, we set the values to \{560, 14, 20, 0.1, 0.1\}; for the HBC and MBA datasets, to \{15, 14, 20, 0.1, 0.1\}; and for the ME and MVC datasets, to \{15, 15, 30, 1, 1\}.

We believe that the choice of LLM is flexible, and various alternatives can be considered. In this study, we select Qwen3-4B \cite{yang2025qwen3} as the LLM used in our experiments due to its ability to run on a single machine with a single GPU while maintaining strong performance.  To reduce training time, we extract and store the biological semantic embeddings in advance, and feed them directly into the model during training.

The prompt we designed for the LLM is as follows:
\begin{quote}
\texttt{System: You are an expert in bioinformatics. Represent the biological state of a cell characterized by the following highly expressed genes. Focus on capturing the functional essence relevant for spatial domain identification.}

\texttt{User: Highly expressed genes: \{gene symbols set\}.}
\end{quote}

To integrate our proposed module with baseline models, we apply it to directly modulate their final latent representations used for clustering during the training phase. The only modification is aligning the dimensionality of the modulation factors with the target latent space. All original architectures and hyperparameters of the baseline models are kept unchanged to ensure a fair comparison.

\begin{figure*}[!htbp]
\centering
\includegraphics[width=0.95\textwidth]{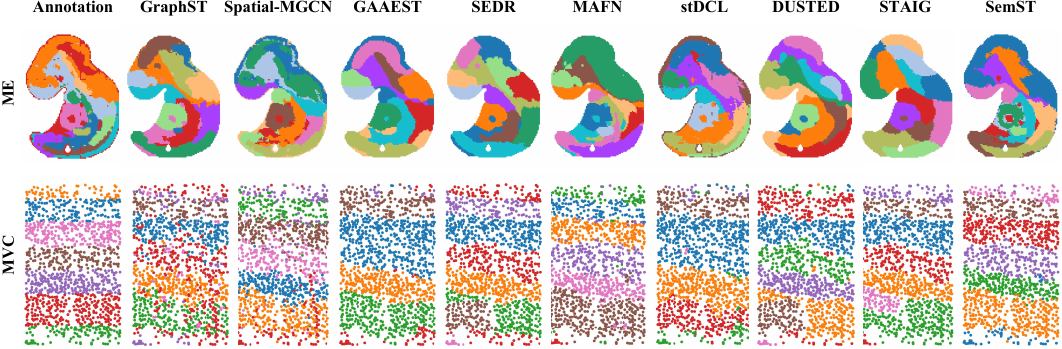}
\caption{Visualization of manual annotations and clustering results produced by SemST and eight other methods on the ME and MVC datasets. Color indicates spatial domains.}
\label{fig:s_v}
\end{figure*}

\begin{figure*}[!htbp]
\centering
\includegraphics[width=0.9\textwidth]{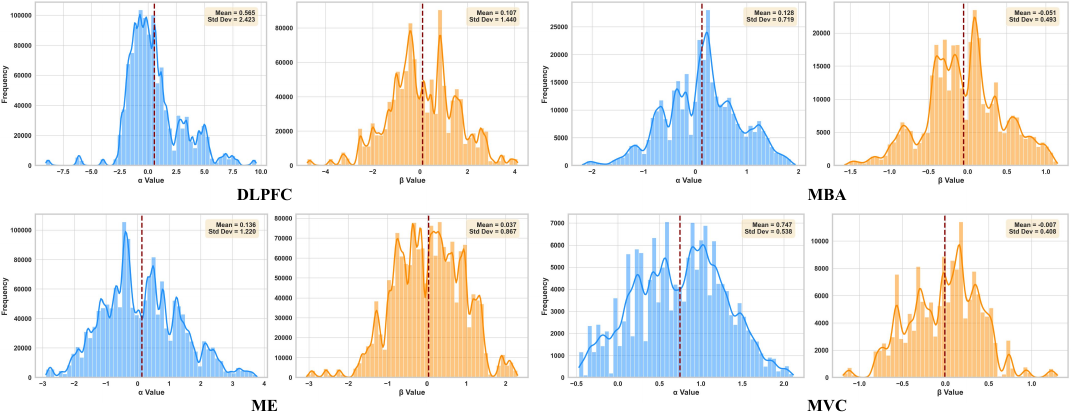}
\caption{Distributions of modulation parameters $\bm{\alpha}$ and $\bm{\beta}$ across the DLPFC slice \#151672, MBA, ME, and MVC datasets.}
\label{fig:bar}
\end{figure*}

\begin{figure*}[!htbp]
\centering
\includegraphics[width=0.9\textwidth]{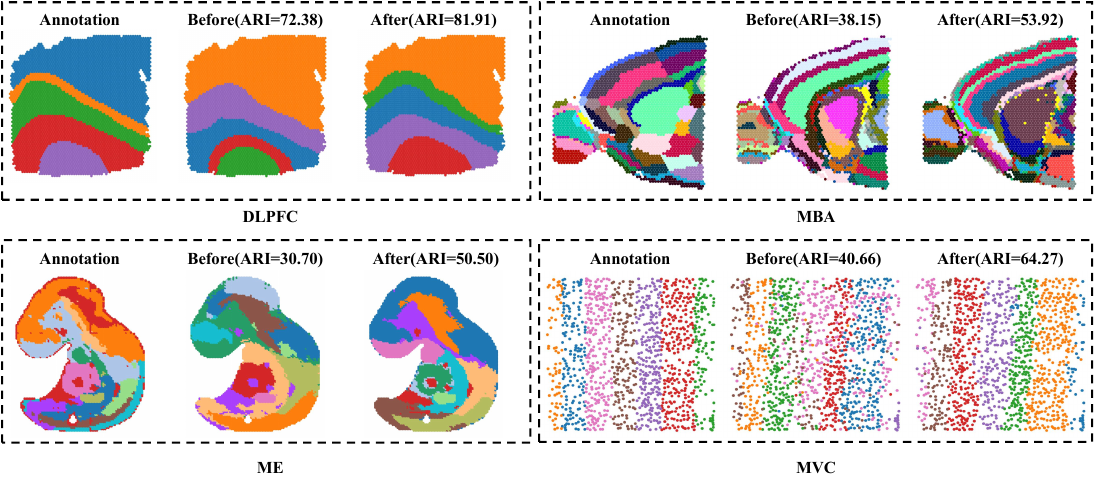}
\caption{Visualization of manual annotations and changes before and after FSM modulation across the DLPFC slice \#151672, MBA, ME, and MVC datasets.}
\label{fig:contrast}
\end{figure*}

\begin{table*}[t]
\centering
\setlength{\tabcolsep}{1mm}
\begin{tabular}{ccllllc}
\toprule
Platform & Spatial resolution & Tissue & Slice & Spots & Genes & clusters \\
\midrule
\multirow{7}{*}{10X Visium} & \multirow{7}{*}{55 $\mu$m} & \multirow{5}{*}{Human 
dorsolateral prefrontal cortex (DLPFC) } & 151508 & 4383 & 33538 & 7 \\
 & & & 151509 & 4789 & 33538 & 7 \\
 & & & 151510 & 4634 & 33538 & 7 \\
 & & & 151671 & 4110 & 33538 & 5 \\
& & & 151672 & 3888 & 33538 & 5 \\
 & & Human breast cancer (HBC) & Section 1 & 3798 & 36601 & 20 \\
  & & Mouse brain anterior (MBA) & Section 1 & 2695 & 32285 & 52 \\
\midrule
Stereo-seq & $\sim$0.5 $\mu$m & Mouse embryo (ME) & E9.5-E1S1 & 5913 & 25568  & 12\\
\midrule
STARmap & $\sim$2 $\mu$m & Mouse visual cortex (MVC) & X & 1207 & 1020 & 7\\
\bottomrule
\end{tabular}
\caption{Description of the datasets used in the study.}
\label{datasets}
\end{table*}

\sisetup{
    detect-all,
    table-align-text-post = false,
    table-format=2.2,
    separate-uncertainty = true
}

\begin{table*}[t]
\centering
\small
\setlength{\tabcolsep}{1.8mm}
\begin{tabular}{
    c| % Datasets 列
    l|
    S[table-format=2.2]@{\,/\,}S[table-format=2.2]@{\,/\,}S[table-format=2.2]
    S[table-format=2.2]@{\,/\,}S[table-format=2.2]@{\,/\,}S[table-format=2.2]
    S[table-format=2.2]@{\,/\,}S[table-format=2.2]@{\,/\,}S[table-format=2.2]
    S[table-format=2.2]@{\,/\,}S[table-format=2.2]@{\,/\,}S[table-format=2.2]
}
\toprule[1pt]
\multirow{2}{*}{Datasets} & \multirow{2}{*}{Methods} 
& \multicolumn{3}{c}{ARI} & \multicolumn{3}{c}{NMI} 
& \multicolumn{3}{c}{ACC} & \multicolumn{3}{c}{F1} \\
\cmidrule(lr){3-5} \cmidrule(lr){6-8} \cmidrule(lr){9-11} \cmidrule(lr){12-14}
& & {Before} & {Concat} & {Add}
& {Before} & {Concat} & {Add}
& {Before} & {Concat} & {Add}
& {Before} & {Concat} & {Add} \\
\midrule
\multirow{9}{*}{HBC} 
& STAGATE {\scriptsize\textcolor{gray}{[Nat. Com.'22]}} 
& {44.66} & {52.89} & \textbf{54.89}
& {67.36} & {67.88} & \textbf{68.22}
& {48.97} & \textbf{55.66} & {52.66}
& {53.50} & \textbf{58.82} & {55.50} \\
& GraphST {\scriptsize\textcolor{gray}{[Nat. Com.'23]}}
& {52.63} & {52.93} & \textbf{55.42}
& {66.96} & {66.82} & \textbf{68.74}
& {54.98} & {53.74} & \textbf{57.69}
& {55.58} & {54.14} & \textbf{58.46} \\
& Spatial-MGCN {\scriptsize\textcolor{gray}{[BIB'23]}}
& \textbf{65.68} & {57.92} & {60.07}
& \textbf{70.83} & {64.90} & {65.40}
& \textbf{64.67} & {59.37} & {59.48}
& \textbf{65.71} & {61.12} & {61.26} \\
& GAAEST {\scriptsize\textcolor{gray}{[Com. Biol.'24]}}
& {52.02} & \textbf{61.21} & {54.81}
& {67.22} & \textbf{69.71} & {67.86}
& {55.71} & \textbf{62.43} & {58.53}
& {55.97} & \textbf{60.69} & {55.83} \\
& SEDR {\scriptsize\textcolor{gray}{[Genome Med.'24]}}
& {43.16} & \textbf{50.84} & {49.10}
& {67.30} & \textbf{69.13} & {67.42}
& {50.34} & \textbf{56.53} & {53.19}
& {54.19} & \textbf{59.95} & {56.64} \\
& MAFN {\scriptsize\textcolor{gray}{[TKDE'24]}}
& {57.49} & \textbf{61.97} & {60.71}
& {62.78} & \textbf{64.82} & {64.58}
& {58.32} & \textbf{61.66} & {60.74}
& {59.55} & {61.91} & \textbf{61.95} \\
& stDCL {\scriptsize\textcolor{gray}{[Adv. Sci.'25]}}
& {55.73} & {51.63} & \textbf{57.89}
& \textbf{70.05} & {66.27} & {68.82}
& {58.19} & {54.48} & \textbf{59.40}
& {56.21} & {53.93} & \textbf{58.25} \\
& DUSTED {\scriptsize\textcolor{gray}{[AAAI'25]}}
& {47.81} & \textbf{50.01} & {43.99}
& \textbf{65.78} & {65.09} & {63.36}
& {48.92} & \textbf{51.00} & {47.58}
& {52.40} & \textbf{53.87} & {51.70} \\
& STAIG {\scriptsize\textcolor{gray}{[Nat. Com.'25]}}
& {57.86} & \textbf{59.24} & {52.19}
& {69.43} & \textbf{69.70} & {66.65}
& {59.82} & \textbf{62.32} & {51.40}
& {63.19} & \textbf{64.69} & {53.68} \\
\midrule

\multirow{9}{*}{MBA} 
& STAGATE {\scriptsize\textcolor{gray}{[Nat. Com.'22]}}
& {35.81} & \textbf{37.30} & {34.27}
& \textbf{72.49} & {71.90} & {71.88} 
& {45.49} & \textbf{45.71} & {44.56}
& {48.80} & \textbf{49.45} & {48.23} \\
& GraphST {\scriptsize\textcolor{gray}{[Nat. Com.'23]}}
& {41.32} & \textbf{42.57} & {42.41}
& \textbf{71.46} & {70.93} & {70.26}
& {47.05} & \textbf{47.31} & {46.31}
& \textbf{48.23} & {46.22} & {45.81} \\
& Spatial-MGCN {\scriptsize\textcolor{gray}{[BIB'23]}}
& {48.34} & {47.85} & \textbf{49.97}
& \textbf{68.03} & {66.94} & {67.93}
& {44.12} & {44.30} & \textbf{46.20}
& {44.87} & {45.57} & \textbf{47.66} \\
& GAAEST {\scriptsize\textcolor{gray}{[Com. Biol.'24]}}
& {43.35} & {45.03} & \textbf{46.33}
& {70.64} & {70.45} & \textbf{71.20}
& \textbf{50.09} & {49.20} & {48.83}
& \textbf{49.29} & {47.13} & {45.68} \\
& SEDR {\scriptsize\textcolor{gray}{[Genome Med.'24]}}
& \textbf{42.05} & {39.74} & {40.40}
& \textbf{71.37} & {70.84} & {71.13}
& {45.71} & \textbf{46.16} & {45.12}
& {45.90} & {48.47} & \textbf{48.48} \\
& MAFN {\scriptsize\textcolor{gray}{[TKDE'24]}}
& {44.15} & {43.56} & \textbf{46.96}
& {67.73} & {66.79} & \textbf{68.87}
& {44.64} & {43.82} & \textbf{46.16}
& {45.48} & {44.16} & \textbf{47.17} \\
& stDCL {\scriptsize\textcolor{gray}{[Adv. Sci.'25]}}
& {42.05} & {39.20} & \textbf{45.35}
& \textbf{70.75} & {69.87} & {70.33}
& \textbf{48.24} & {45.19} & {47.20}
& \textbf{45.06} & {43.74} & {42.85} \\
& DUSTED {\scriptsize\textcolor{gray}{[AAAI'25]}}
& \textbf{35.86} & {32.89} & {31.65}
& \textbf{71.13} & {70.80} & {70.13}
& \textbf{43.75} & {42.08} & {40.96}
& \textbf{47.13} & {46.15} & {44.17} \\
& STAIG {\scriptsize\textcolor{gray}{[Nat. Com.'25]}} 
& {33.45} & \textbf{34.61} & {32.67}
& {70.61} & \textbf{70.88} & {70.86}
& \textbf{44.12} & {43.75} & {43.78}
& \textbf{46.68} & {45.66} & {44.92} \\
\bottomrule[1pt]
\end{tabular}
\caption{A comparison of clustering performance between original baseline models (Before) and versions enhanced by an LLM-derived embedding via concatenation (Concat) or addition (Add). \textbf{Bold} values highlight the best result.}
\label{tab:ablation-cat-add}
\end{table*}

\section{More Experiment Results}
Figure \ref{fig:s_v} presents the visualizations of manual annotations and clustering results obtained by SemST and eight baseline methods on the ME and MVC datasets. The spatial domains identified by SemST show greater consistency with the manual annotations.

Table 2 details the performance of two naive fusion methods (concatenation and addition) for incorporating LLM-derived semantic embeddings into the existing methods. Although these embeddings, which encode biological knowledge, generally augment the spatial representations, their simple integration fails to yield consistent gains in clustering performance and occasionally results in performance degradation. This finding motivates the development of a more refined mechanism for a synergistic integration of semantic and spatial information.

To gain insight into the behavior of the FSM module, we visualize the distributions of the learned scaling ($\bm{\alpha}$) and shifting ($\bm{\beta}$) parameters across four datasets, as shown in Figure \ref{fig:bar}. The distributions consistently exhibit non-zero means and significant variance, indicating that the module learns a diverse set of non-trivial modulations. Figure \ref{fig:contrast} visualizes the modulation performance of the FSM module across four datasets. The method effectively refines the initial clustering ("Before") to align with the ground truth ("Annotation"), yielding more accurate spatial domains ("After").

\section{Discussion on Gene Texts}
\begin{figure}[t]
\centering
\includegraphics[width=0.98\columnwidth]{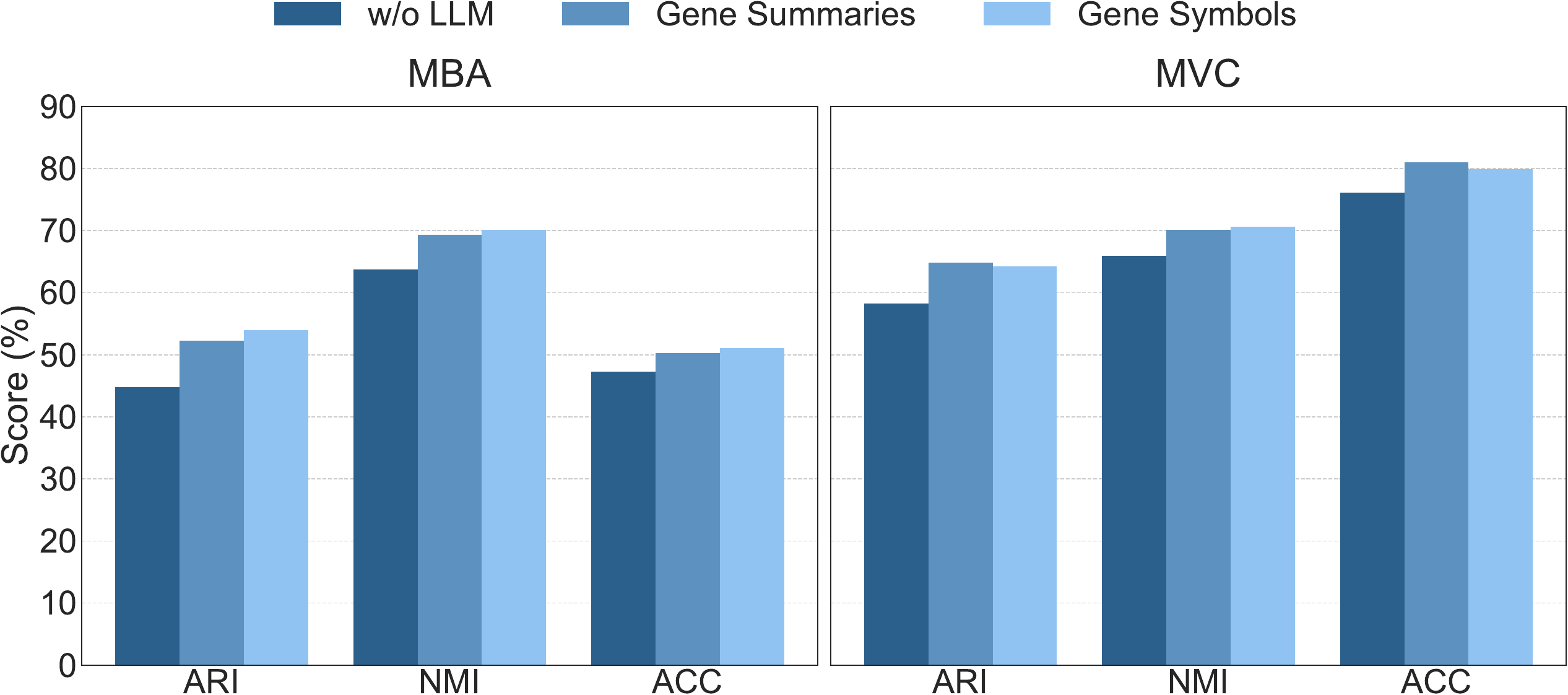}
\caption{Comparison of clustering performance using Gene Summaries and Gene Symbols as input, and the baseline method (w/o LLM).}
\label{bar_c}
\end{figure}

We also explored an alternative approach for generating semantic embeddings. For each gene, we obtained its semantic embedding by inputting the gene along with its NCBI summary into the Qwen3-Embedding-4B model. The semantic embedding for each spot was then calculated by average-pooling the embeddings of its highly expressed genes. As shown in Figure \ref{bar_c}, the resulting clustering performance is very similar to that of the proposed method. Intuitively, the summary-based method has the advantage of reducing the likelihood of LLM hallucinations, whereas the gene symbol-based method is simpler and better at capturing synergistic interactions and holistic functions among genes. Overall, these results collectively highlight that the integration of gene-level biological information can facilitate deeper insights into spatial transcriptomic data.

\end{document}